\pdfoutput=1

\documentclass[11pt]{article}

\usepackage{naacl2021}

\usepackage{times}
\usepackage{latexsym}

\usepackage[T1]{fontenc}

\usepackage[utf8]{inputenc}

\usepackage{microtype}

\usepackage{soul}
\usepackage{url}
\usepackage{hyperref}
\usepackage[utf8]{inputenc}
\usepackage{caption}
\usepackage{graphicx}
\usepackage{amsmath,amsfonts,amssymb}
\usepackage{booktabs}
\usepackage{subcaption}
\usepackage{mathtools}
\usepackage{makecell}

\usepackage{tabu}

\usepackage{enumitem}
\usepackage{siunitx}
\usepackage{multirow}
\usepackage{algorithm}
\usepackage{setspace}
\usepackage{algcompatible}
\usepackage[noend]{algpseudocode}
\makeatletter
\def\BState{\State\hskip-\ALG@thistlm}
\makeatother

\algdef{SE}[SUBALG]{Indent}{EndIndent}{}{\algorithmicend\ }%
\algtext*{Indent}
\algtext*{EndIndent}

\newcommand{\parm}{\mathord{\color{black!33}\bullet}}

\newcommand\METHOD{VID~}

\title{View Distillation with Unlabeled Data for Extracting Adverse Drug Effects from User-Generated Data}

\author{

Payam Karisani \\
Emory University \\
\texttt{pkarisa@emory.edu} \\\And

Jinho D. Choi \\
Emory University \\
\texttt{jinho.choi@emory.edu} \\ \And

Li Xiong \\
Emory University \\
\texttt{lxiong@emory.edu} \\

  }

\begin{document}
\maketitle
\begin{abstract}

We present an algorithm based on multi-layer transformers for identifying Adverse Drug Reactions (ADR) in social media data. Our model relies on the properties of the problem and the characteristics of contextual word embeddings to extract two views from documents. Then a classifier is trained on each view to label a set of unlabeled documents to be used as an initializer for a new classifier in the other view. Finally, the initialized classifier in each view is further trained using the initial training examples. We evaluated our model in the largest publicly available ADR dataset. The experiments testify that our model significantly outperforms the transformer-based models pretrained on domain-specific data.

\end{abstract}

\vspace{-0.2cm}
\section{Introduction} \label{sec:intro}
\vspace{-0.1cm}

Social media has made substantial amount of data available for various applications in the financial, educational, and health domains. Among these, the applications in healthcare have a particular importance. Although previous studies have demonstrated that the self-reported online social data is subject to various biases \cite{olteanu2016social}, this data has enabled many applications in the health domain, including tracking the spread of influenza \cite{aramaki2011twitter}, detecting the reports of the novel coronavirus \cite{our-corona}, and identifying various illness reports \cite{wespad}.

One of the well-studied areas in online public health monitoring is the extraction of adverse drug reactions (ADR) from social media data. ADRs are the unintended effects of drugs for prevention, diagnosis, or treatment. The researchers in \citet{adr-time} reported that consumers, on average, report the negative effect of drugs on social media 11 months earlier than other platforms. This highlights the importance of this task. Another team of researchers in \citet{adr-review} reviewed more than 50 studies and reported that the prevalence of ADRs across multiple platforms ranges between 0.2\% and 8.0\%, which justifies the difficulty of this task. In fact, despite the long history of this task in the research community \cite{adr-first}, for various reasons, the performance of the state-of-the-art models is still unsatisfactory. Social media documents are typically short and their language is informal \cite{filter-expansion}. Additionally, the imbalanced class distributions in ADR task has exacerbated the problem.

In this study we propose a novel model for extracting ADRs from Twitter data. Our model which we call View Distillation (\METHOD\!) relies on the existence of two views in the tweets that mention drug names. We use unlabeled data to transfer the knowledge from the classifier in each view to the classifier in the other view. Additionally, we use a finetuning technique to mitigate the impact of noisy pseudo-labels after the initialization \cite{self-pretraining}. As straightforward as it is to implement, our model achieves the state-of-the-art performance in the largest publicly available ADR dataset, i.e., SMM4H dataset. Our contributions are as follows: 1) We propose a novel algorithm to transfer knowledge across models in multi-view settings, 2) We propose a new technique to efficiently exploit unlabeled data in the supervised ADR task, 3) We evaluate our model in the largest publicly available ADR dataset, and show that it yields an additive improvement to the common practice of language model pretraining in this task. To our knowledge, our work is the first study that reports such an achievement. Next, we provide a brief overview of the related studies.

\vspace{-0.2cm}
\section{Related Work} \label{sec:rel-work}
\vspace{-0.1cm}

Researchers have extensively explored the applications of ML and NLP models in extracting ADRs from user-generated data. Perhaps one of the early reports in this regard is published in \citet{adr-first}, where the authors utilize the related lexicons and extraction patterns to identify ADRs in user reviews. With the surge of neural networks in text processing, subsequently, the traditional models were aggregated with these techniques to achieve better generalization \cite{adr-crf-rnn}. The recent methods for extracting ADRs entirely rely on neural network models, particularly on multi-layer transformers \cite{attention}. 

In the shared task of SMM4H 2019 \cite{smm4h-2019}, the top performing run was BERT model \cite{bert} pretrained on drug related tweets. Remarkably, one year later in the shared task of SMM4H 2020 \cite{smm4h-2020}, again a variant of pretrained BERT achieved the best performance \cite{roberta}. Here, we propose an algorithm to improve on pretrained BERT in this task. Our model relies on multi-view learning and exploits unlabeled data. To our knowledge, our model is the first approach that improves on the domain-specific pretrained BERT.

\vspace{-0.2cm}
\section{Proposed Method} \label{sec:method}
\vspace{-0.1cm}

Our model for extracting the reports of adverse drug effects rely on the properties of contextual neural word embeddings. Previous research on Word Sense Disambiguation (WSD) \cite{embed-wsd} has demonstrated that contextual word embeddings can effectively encode the context in which words are used. Although the representations of the words in a sentence are assumed to be distinct, they still possess shared characteristics. This is justified by the observation that the techniques such as self-attention \cite{attention}, which a category of contextual word embeddings employ \cite{bert}, rely on the interconnected relations between word representations.

\begin{figure}
\centering
\includegraphics[width=3.1in]{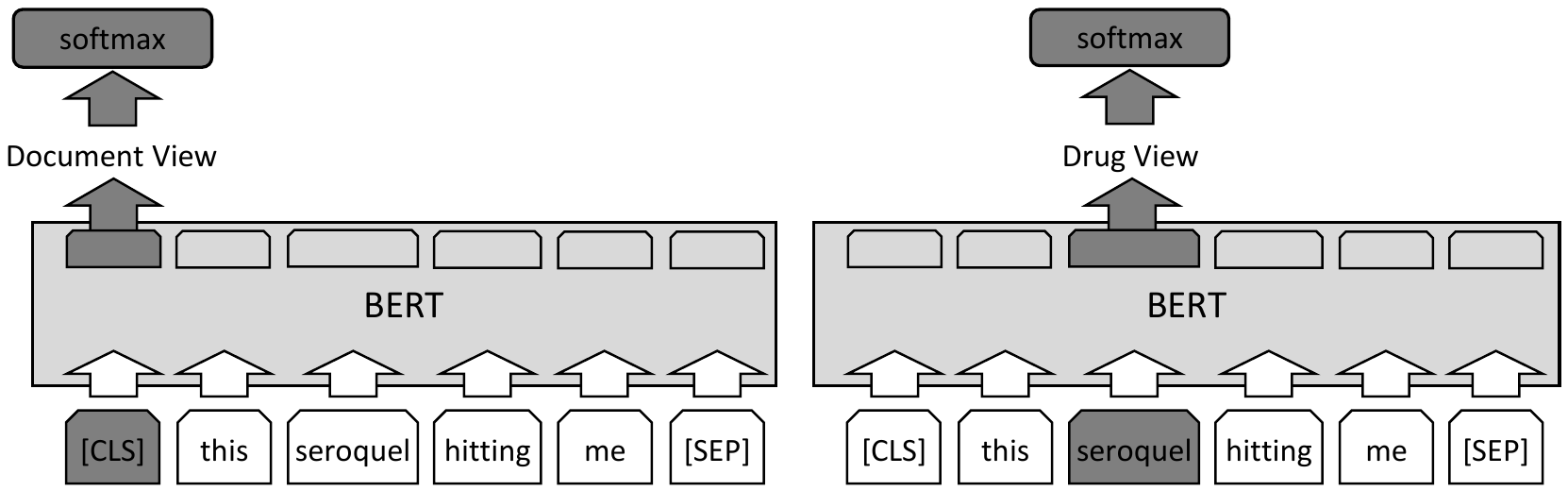}
\caption{The illustration of the document and drug views in our model. We have used BERT as an encoder. See \citet{bert} for the format of input tokens.} \label{fig:diagram}
\vspace{-0.5cm}
\end{figure}

This property is particularly appealing when documents are short, therefore, word representations, if are adjusted accordingly, can be exploited to extract multiple representations for a single document. In fact, previous studies have demonstrated that word contexts can be used to process short documents, e.g., see the models proposed in \citet{sent-cotest} and \citet{co-decomp} for event extraction using hand-crafted features and contextual word embeddings respectively. Therefore, we use the word representations of drug mentions in user postings as the secondary view along the document representations of user postings in our model. As a concrete example, from the hypothetical tweet \textit{``this seroquel hitting me''}, we extract one representation from the entire document and another representation from the drug name\footnote{We assume every user posting contains only one drug name, in cases that there are multiple names we can use the first occurrence.} Seroquel. In continue, we call these two views the document and drug views. Figure \ref{fig:diagram} illustrates these two views using BERT \cite{bert} as an encoder.

Given the two views we can either concatenate the two sets of features and train a classifier on the resulting feature vector or use a co-training framework as described in \citet{co-decomp}. However, the former is not exploiting the abundant amount of unlabeled data, and the latter is resource intensive, because it is iterative, and also it has shown to be effective only in semi-supervised settings where there are only a few hundred training examples available. Therefore, below we propose an approach to effectively use the two views along the available unlabeled data in a supervised setting.

In the first step, we assume the classifier in each view is a student model and train this classifier using the pseudo-labels generated by the counterpart classifier. Since the labeled documents are already annotated, we carry out this step using the unlabeled documents. More concretely, let $L$ and $U$ be the sets of labeled and unlabeled user postings respectively. Moreover, let $L_{d}$ and $L_{g}$ be the sets of representations extracted from the document and drug views of the training examples in the set $L$; and let $U_{d}$ and $U_{g}$ be the document and drug representations of the training examples in the set $U$. To carry out this step, we train a classifier $C_{d}$ on the representations in $L_{d}$ and probabilistically, with temperature $T$ in the softmax layer, label the representations in $U_{d}$. Then we use the association between the representations in $U_{d}$ and $U_{g}$ to construct a pseudo-labeled dataset of $U_{g}$. This dataset along its set of probabilistic pseudo-labels is used in a distillation technique \cite{distill} to train a classifier called $\widehat{C_{g}}$. Correspondingly, we use the set $L_{g}$ to train a classifier $C_{g}$, then label the set $U_{g}$ and use the association between the data points in $U_{g}$ and $U_{d}$ to construct a pseudo-labeled dataset in the document view to train the classifier~$\widehat{C_{d}}$.

The procedure above results in two classifiers $\widehat{C_{d}}$ and $\widehat{C_{g}}$. The classifier in each view is \textit{initialized} by the knowledge transferred from the other view. However, the pseudo-labels that are used to train each classifier can be noisy. Thus, in order to reduce the negative impact of this noise, in the next step, we use the training examples in the sets $L_{d}$ and $L_{g}$ to further finetune these two classifiers respectively. To finetune $\widehat{C_{d}}$ we use the objective function below:
\begin{equation} \label{eq:loss-d}
\small
\setlength{\jot}{0pt}
\setlength{\abovedisplayskip}{0pt}
\setlength{\belowdisplayskip}{0pt}
\medmuskip=0mu
\thinmuskip=0mu
\thickmuskip=0mu
\nulldelimiterspace=0pt
\scriptspace=0pt
\begin{split}
\mathcal{L}_{d}= \frac{1}{\left|L_{d}\right|}\sum_{v \in L_{d}}^{} (1 - \lambda) J(\widehat{C_{d}}(v), y_{v}) + \lambda J(\widehat{C_{d}}(v), C_{d}(v)),
\end{split}
\end{equation}
where $J$ is the cross-entropy loss, $y_v$ is the ground-truth label of the training example $v$, and $\lambda$ is a hyper-parameter to govern the impact of the two terms in the summation. The first term in the summation, is the regular cross-entropy between the output of $\widehat{C_{d}}$ and the ground-truth labels. The second term is the cross-entropy between the outputs of $\widehat{C_{d}}$ and $C_{d}$. We use the output of $C_{d}$ as a regularizer to train $\widehat{C_{d}}$ in order to increase the entropy of this classifier for the prediction phase. Previous studies have shown that penalizing low entropy predictions increases generalization \cite{entropy-general}. We argue that this is particularly important in the ADR task, where the data is highly imbalanced. Note that, even though $C_{d}$ is trained on the training examples in $L_{d}$, the output of this classifier for the training examples is not sparse--particularly for the examples with uncommon characteristics. Thus, we use these soft-labels\footnote{Again, we use temperature $T$ in the softmax layer to train using the soft-labels.} along the ground-truth labels to train $\widehat{C_{d}}$. Respectively, we use the objective function below to finetune $\widehat{C_{g}}$:
\begin{equation} \label{eq:loss-g}
\small
\setlength{\jot}{0pt}
\setlength{\abovedisplayskip}{0pt}
\setlength{\belowdisplayskip}{0pt}
\medmuskip=0mu
\thinmuskip=0mu
\thickmuskip=0mu
\nulldelimiterspace=0pt
\scriptspace=0pt
\begin{split}
\mathcal{L}_{g}= \frac{1}{\left|L_{g}\right|}\sum_{v \in L_{g}}^{} (1 - \lambda) J(\widehat{C_{g}}(v), y_{v}) + \lambda J(\widehat{C_{g}}(v), C_{g}(v)),
\end{split}
\end{equation}
where the notation is similar to that of Equation \ref{eq:loss-d}. Here, we again use the output of $C_{g}$ as a regularizer to train $\widehat{C_{g}}$. In the evaluation phase, to label the unseen examples, we take the average of the outputs of the two classifiers $\widehat{C_{d}}$ and $\widehat{C_{g}}$.

\begin{algorithm}
\footnotesize
\caption{Overview of \METHOD}\label{alg:summary}
\begin{algorithmic}[1]
\algrenewcommand\algorithmicindent{7pt}
\Procedure{VID}{}
    \State \textbf{Given:}
    \Indent
        \State $L: \text{Set of labeled documents}$ 
        \State $U: \text{Set of unlabeled documents}$
    \EndIndent
    \State \textbf{Return:}
    \Indent
        \State $\text{Two classifiers}~\widehat{C_{d}}~\text{and}~\widehat{C_{g}}$
    \EndIndent
    \State \textbf{Execute:}
    \Indent
        \State Derive two sets of representations $L_{d}$ and $L_{g}$ from $L$ \label{alg-line:derive-l}
        \State Derive two sets of representations $U_{d}$ and $U_{g}$ from $U$ \label{alg-line:derive-g}
        \State Use $L_{d}$ to train classifier $C_{d}$ \label{alg-line:train-d}
        \State Use $L_{g}$ to train classifier $C_{g}$ \label{alg-line:train-g}
        \State Use $C_{d}$ to probabilistically label $U_{d}$ \label{alg-line:g-0}
        \State Transfer labels of $U_{d}$ to $U_{g}$ and use them to train $\widehat{C_{g}}$
        \State Finetune $\widehat{C_{g}}$ using Equation \ref{eq:loss-g} \label{alg-line:g-1}
        \State Use $C_{g}$ to probabilistically label $U_{g}$ \label{alg-line:d-0}
        \State Transfer labels of $U_{g}$ to $U_{d}$ and use them to train $\widehat{C_{d}}$
        \State Finetune $\widehat{C_{d}}$ using Equation \ref{eq:loss-d} \label{alg-line:d-1}
        \State \textbf{Return} $\widehat{C_{d}}~\text{and}~\widehat{C_{g}}$
    \EndIndent
\EndProcedure
\end{algorithmic}
\end{algorithm}

Algorithm \ref{alg:summary} illustrates our model (\METHOD\!) in Structured English. On Lines \ref{alg-line:derive-l} and \ref{alg-line:derive-g} we derive the document and drug representations from the sets $L$ and $U$. On Lines \ref{alg-line:train-d} and \ref{alg-line:train-g} we use the labeled training examples in the two views to train $C_{d}$ and $C_{g}$. On Lines \ref{alg-line:g-0}-\ref{alg-line:g-1} we train and finetune $\widehat{C_{g}}$, and on Lines \ref{alg-line:d-0}-\ref{alg-line:d-1} we train and finetune $\widehat{C_{d}}$. Finally, we return $\widehat{C_{d}}$ and $\widehat{C_{g}}$. In the next section, we describe our experimental setup.

\vspace{-0.2cm}
\section{Experimental Setup} \label{sec:setup}
\vspace{-0.1cm}

We evaluated our model in the largest publicly available ADR dataset, i.e., the SMM4H dataset. This dataset consists of 30,174 tweets. The training set in this dataset consists of 25,616 tweets of which 9.2\% are positive. The labels of the test set are not publicly available. The evaluation in the dataset must be done via the CodaLab website. We compare our model with two sets of baselines: 1) a set of baselines that we implemented, 2) the set of baselines that are available on the CodaLab website\footnote{Available at: \mbox{\href{https://competitions.codalab.org/competitions/20798}{\nolinkurl{https://competitions.codalab.org/SMM4H}}}. The 2020 edition of the shared task is not online anymore. Therefore, for a fair comparison with the baselines, we do not use RoBERTa in our model, and instead use pre-trained BERT model.}. 

Our own baseline models are: \textbf{BERT}, the base variant of the pretrained BERT model \cite{bert}, as published by Google. \textbf{BERT-D}, a domain-specific pretrained BERT model. This model is similar to the previous baseline, however, it is further pretrained on 800K unlabeled drug-related tweets that we collected from Twitter. We pretrained this model for 6 epochs using the next sentence prediction and the masked language model tasks. \textbf{BERT-D-BL}, a bi-directional LSTM model. In this model we used BERT-D followed by a bi-directional LSTM network \cite{lstm}.

We also compare our model with all the baselines available on the CodaLab webpage. These baselines include published and unpublished models. They also cover models that purely rely on machine learning models and those that heavily employ medical resources; see \citet{smm4h-2019} for the summary of a subset of these models.

We used the Pytorch implementation of BERT \cite{bert-impl}. we used two instances of BERT-D as the classifiers in our model--see Figure \ref{fig:diagram}. Please note that using domain-specific pretrained BERT in our framework makes any improvement very difficult, because the improvement in the performance should be additive. We used the training set of the dataset to tune for our two hyper-parameters $T$ and $\lambda$. The optimal values of these two hyper-parameters are 2 and 0.5 respectively. We trained all the models for 5 epochs\footnote{We used 20\% of the training set for validation, and observed that the models overfit if we train more than 5 epochs.}. During the tuning, we observed that the finetuning stage in our model requires much fewer training steps, therefore, we finetuned for only 1 epoch. In our model, we used the same set of unlabeled tweets that we used to pretrain BERT-D. This verifies that, indeed, our model extracts new information that cannot be extracted using the regular language model pretraining. As required by SMM4H we tuned for F1 measure. In the next section, we report the F1, Precision, and Recall metrics.

\vspace{-0.2cm}
\section{Results and Analysis} \label{sec:result}
\vspace{-0.1cm}

\begin{table}
\centering
\small
\begin{tabu}{p{0.7in} p{0.7in} p{0.3in} p{0.4in} p{0.4in} } \Xhline{3\arrayrulewidth}
\multicolumn{1}{c}{\textbf{Type}} & \textbf{Method} & \textbf{F1} & \textbf{Precision} & \textbf{Recall} \\ \Xhline{3\arrayrulewidth}
\multicolumn{1}{c}{\multirow{3}{*}{Our Impl.}} & BERT & 0.57 & 0.669 & 0.50 \\
\multicolumn{1}{c}{} & BERT-D & 0.62 & 0.736 & 0.54 \\ 
\multicolumn{1}{c}{} & BERT-D-BL & 0.61 & \textbf{0.749} & 0.52 \\ \hline
\multicolumn{1}{c}{\multirow{3}{*}{CodaLab}} & Sarthak & 0.65 & 0.661 & 0.65 \\
\multicolumn{1}{c}{} & leebean337 & 0.67 & 0.600 & \textbf{0.76} \\ 
\multicolumn{1}{c}{} & aab213 & 0.67 & 0.608 & 0.75 \\ \hline
\multicolumn{1}{c}{} & \METHOD & \textbf{0.70} & 0.678 & 0.72 \\ 
\Xhline{3\arrayrulewidth}
\end{tabu}
\caption{F1, Precision, and Recall of our model (\METHOD\!) in comparison with the baselines.} \label{tbl:results}
\vspace{-0.5cm}
\end{table}

Table \ref{tbl:results} reports the performance of our model in comparison with the baseline models--only the top three CodaLab baselines are listed here. We see that our model significantly outperforms all the baseline models. We also observe that the performances of our implemented baseline models are lower than that of the CodaLab models. This difference is mainly due to the gap between the size of the unlabeled sets for the language model pretraining in the experiments--ours is 800K, but the top CodaLab model used a corpus of 1.5M examples. This suggests that our model can potentially achieve a better performance if there is a larger unlabeled corpus available. 

\begin{table}
\centering
\small
\begin{tabu}{p{1.5in}  p{0.3in} p{0.5in} p{0.4in} } \Xhline{3\arrayrulewidth}
\multicolumn{1}{c}{\textbf{Method}} & \textbf{F1} & \textbf{Precision} & \textbf{Recall} \\ \Xhline{3\arrayrulewidth}
\multicolumn{1}{c}{\textit{Document-View}} & 0.62 & 0.736 & 0.54 \\ 
\multicolumn{1}{c}{\textit{Drug-View}} & 0.63 & 0.706 & 0.570 \\ 
\multicolumn{1}{c}{\textit{Combined-View}} & 0.63 & 0.745 & 0.543 \\ \hline
\multicolumn{1}{c}{\textit{\METHOD}} & 0.70 & 0.678 & 0.72 \\ \Xhline{3\arrayrulewidth}
\end{tabu}
\caption{F1, Precision, and Recall of \METHOD in comparison to the performance of the classifiers trained on the document, drug, and combined views.} \label{tbl:individual-cls}
\vspace{-0.2cm}
\end{table}

\begin{table}
\centering
\small
\begin{tabu}{p{1.5in}  p{0.3in} p{0.5in} p{0.4in} } \Xhline{3\arrayrulewidth}
\multicolumn{1}{c}{\textbf{Method}} & \textbf{F1} & \textbf{Precision} & \textbf{Recall} \\ \Xhline{3\arrayrulewidth}
\multicolumn{1}{c}{\textit{P-Doc-F-Doc}} & 0.69 & 0.658 & 0.71 \\ 
\multicolumn{1}{c}{\textit{P-Drug-F-Drug}} & 0.68 & 0.681 & 0.68 \\ 
\multicolumn{1}{c}{\textit{P-Doc-F-Drug}} & 0.70 & 0.674 & 0.72 \\ 
\multicolumn{1}{c}{\textit{P-Drug-F-Doc}} & 0.69 & 0.655 & 0.72 \\ \hline
\multicolumn{1}{c}{\textit{\METHOD}} & 0.70 & 0.678 & 0.72 \\ \Xhline{3\arrayrulewidth}
\end{tabu}
\caption{Performance of \METHOD in comparison to the performance of the classifiers pretrained on the document or drug pseudo-labels (indicated by P-$\{\parm\}$) and finetuned on the document or drug training examples (indicated by F-$\{\parm\}$).} \label{tbl:pretrain}
\vspace{-0.5cm}
\end{table}

Table \ref{tbl:individual-cls} reports the performance of \METHOD in comparison to the classifiers trained on the document and drug representations. We also concatenated the two representations and trained a classifier on the resulting feature vector, denoted by \textit{Combined-View}. We see that our model substantially outperforms all three models. Table \ref{tbl:pretrain} compares our model with the classifiers with different pretraining and finetuning resources. Again, we see that \METHOD is comparable to the best of these models. We also observe 2 percent absolute improvement by comparing \textit{P-Drug-F-Drug} and \textit{P-Doc-F-Drug}, which signifies the efficacy of View Distillation.

In summary, we evaluated our model in the largest publicly available ADR dataset and compared with the state-of-the-art baseline models that use domain specific language model pretraining. We showed that our model outperforms these models, even though it uses a smaller unlabeled corpus. We also carried out a set of experiments and demonstrated the efficacy of our proposed techniques.

\vspace{-0.2cm}
\section{Conclusions} \label{sec:conclusion}
\vspace{-0.1cm}

In this study we proposed a novel model for extracting adverse drug effects from user generated content. Our model relies on unlabeled data and a novel technique called view distillation. We evaluated our model in the largest publicly available ADR dataset, and showed that it outperforms the existing BERT-based models.


\bibliography{custom}
\bibliographystyle{acl_natbib}



\end{document}